\documentclass{article}
\pdfoutput=1

\usepackage[final, nonatbib]{neurips_2024}

\usepackage[utf8]{inputenc} 
\usepackage[T1]{fontenc}    
\usepackage{hyperref}       
\usepackage{url}            
\usepackage{booktabs}       
\usepackage{amsfonts}       
\usepackage{nicefrac}       
\usepackage{microtype}      
\usepackage{xcolor}         
\usepackage{bbm}
\usepackage{svg}
\usepackage{pifont}
\usepackage{graphicx} 
\usepackage{grffile}

\usepackage{booktabs,siunitx}
\usepackage{caption}
\usepackage{subcaption}
\usepackage{enumitem}
\usepackage{algorithm}

\usepackage{tikz}

\usepackage{filecontents}

\usepackage{listings}
\usepackage{xcolor}
\usepackage{footnote}

\usepackage{adjustbox}
\usepackage{multirow}
\usepackage{colortbl}
\usepackage{float}
\usepackage{algorithmicx}   
\usepackage{algpseudocode}
\usepackage{setspace}
\usepackage{makecell}
\usepackage{wrapfig}

\usepackage{amsfonts}

\usepackage{tablefootnote}
\usepackage{threeparttable}

\usepackage{amsmath}
\DeclareMathOperator*{\argmax}{argmax}
\definecolor{LightCyan}{rgb}{0.88,1,1}

\newcommand{\putname}{\textit{(FL)$^2$}}

\hypersetup{
    colorlinks=true,
    urlcolor=magenta
}

\title{(FL)${}^2$: Overcoming Few Labels in Federated Semi-Supervised Learning} 

\author{%
  Seungjoo Lee \hfill Thanh-Long V. Le \hfill Jaemin Shin \hfill Sung-Ju Lee \\ [3pt]
  KAIST\\ [2pt]
  Republic of Korea \\ [2pt]
  \texttt{\{seungjoo.lee,thanhlong0780,jaemin.shin,profsj\}@kaist.ac.kr}
}

\begin{document}

\maketitle

\begin{abstract}
Federated Learning (FL) is a distributed machine learning framework that trains accurate global models while preserving clients' privacy-sensitive data. However, most FL approaches assume that clients possess labeled data, which is often not the case in practice. Federated Semi-Supervised Learning~(FSSL) addresses this label deficiency problem, targeting situations where only the server has a small amount of labeled data while clients do not. However, a significant performance gap exists between Centralized Semi-Supervised Learning~(SSL) and FSSL. This gap arises from confirmation bias, which is more pronounced in FSSL due to multiple local training epochs and the separation of labeled and unlabeled data. We propose \putname{}, a robust training method for unlabeled clients using \textit{sharpness-aware consistency regularization}. We show that regularizing the original pseudo-labeling loss is suboptimal, and hence we carefully select unlabeled samples for regularization. We further introduce \textit{client-specific adaptive thresholding} and \textit{learning status-aware aggregation} to adjust the training process based on the learning progress of each client. Our experiments on three benchmark datasets demonstrate that our approach significantly improves performance and bridges the gap with SSL, particularly in scenarios with scarce labeled data. The source code is available at \texttt{\href{https://github.com/seungjoo-ai/FLFL-NeurIPS24}{https://github.com/seungjoo-ai/FLFL-NeurIPS24}}
\end{abstract}
\section{Introduction}

Federated learning~(FL)~\cite{fedavg} is a distributed machine learning system that trains accurate global models while preserving clients' privacy-sensitive data. Each FL client trains its local model on their device using only their data, and the server aggregates these local models into a global model. As a result, clients' private data is protected as only the local models' weights are shared with the server.

Because of its privacy-preserving nature, FL has garnered recent attention, with efforts to make it reliable and efficient~\cite{fedbalancer,lai2021oort, sattler2019robust}. However, most previous FL studies assumed that clients have labeled data, which is unrealistic in practical settings for two reasons. First, clients are often reluctant or lack the motivation to label data. Second, certain data types require domain expertise during the labeling process~\cite{yang2021federated, elbir2022federated}. For example, labeling medical data demands specialized knowledge and expertise. Similarly, sensory data, which can have multiple dimensions,  is difficult for most clients to interpret accurately. Therefore, we envision a \textit{labels-at-server}~\cite{jeong2020federated} scenario as more realistic for FL, where a small amount of labeled data is available only at the server while the clients' data remains unlabeled.

\begin{figure}[t]
    \centering
    \includegraphics[width=0.7\textwidth]{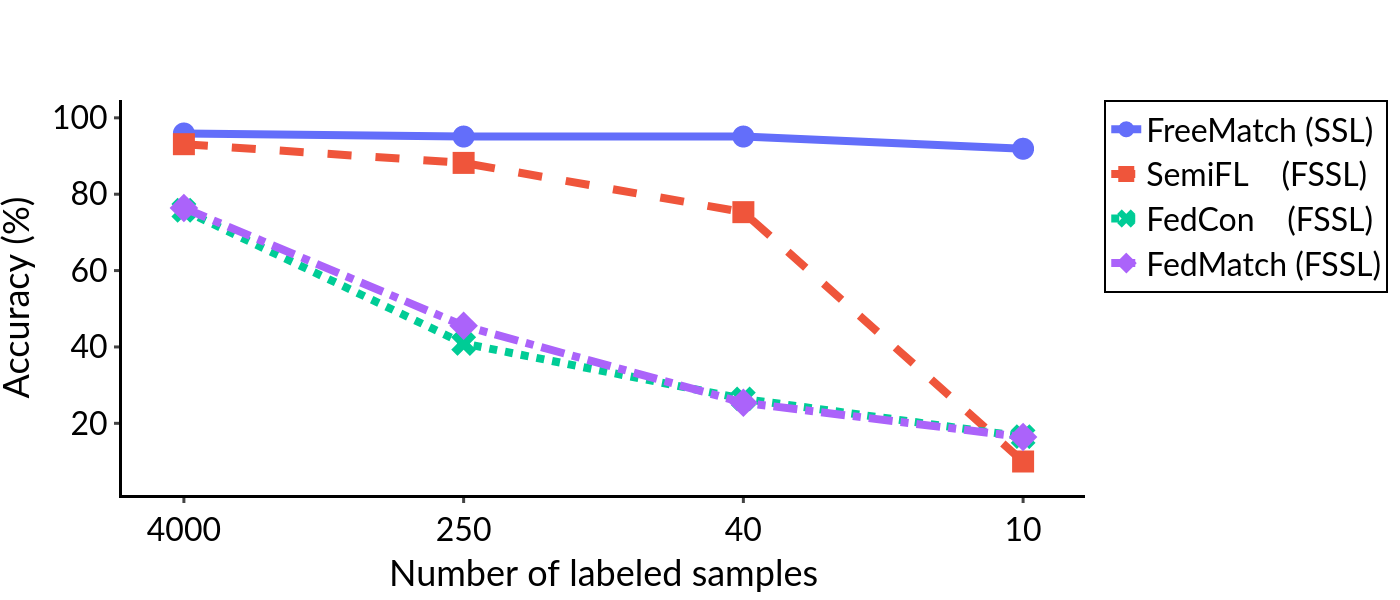}
    \caption{ Comparison of SSL and FSSL algorithms on CIFAR-10 with varying numbers of labeled samples, where FreeMatch~\cite{wang2022freematch} represents SSL, and SemiFL~\cite{diao2022semifl}, FedCon~\cite{long2021fedcon}, and FedMatch~\cite{jeong2020federated} represent FSSL.}
    \label{fig:motivation}
    \vspace{-1em}
\end{figure}

Various Federated Semi-Supervised Learning (FSSL) approaches~\cite{diao2022semifl, jeong2020federated, long2021fedcon, zhang2021ssfl, kim2022fedrgd} have been developed for the \textit{labels-at-server} scenario.
However, there is a substantial performance gap between FSSL and centralized Semi-Supervised Learning~(SSL), particularly when labeled data is limited. Fig.~\ref{fig:motivation} illustrates this issue across varying amounts of labeled data on the CIFAR10 dataset~\cite{alex2009learningcifar}. When a sufficient amount of labeled data is available, the performance difference between SSL and FSSL is minimal. However, this gap widens considerably as the quantity of labeled data decreases.

We point out \textit{confirmation bias} as the primary cause, where the model tends to overfit to easy-to-learn samples or incorrectly pseudo-labeled data~\cite{nguyen2023boostingrefix}. This issue is particularly pronounced in FSSL as the training process involves multiple local epochs on clients~\cite{diao2022semifl, qiu2024fedanchor, lee2023fedsol}. This extended training on localized data accelerates the overfitting process, making the model more susceptible to confirmation bias. Moreover, labeled and unlabeled data are kept separate in a \textit{labels-at-server} setting. Unlike in centralized SSL where labeled and unlabeled objectives could be jointly optimized, this separation in FSSL prevents effective co-optimization, further contributing to the performance gap.

We propose \textbf{F}ew-\textbf{L}abels \textbf{F}ederated semi-supervised \textbf{L}earning, abbreviated as \putname{}, to mitigate \textit{confirmation bias} in FSSL using (1)~\textit{client-specific adaptive thresholding}, (2)~\textit{sharpness-aware consistency regularization}, and (3)~\textit{learning status-aware aggregation}. Previous FSSL approaches~\cite{jeong2020federated, diao2022semifl} use a fixed threshold to obtain high-confidence pseudo-labels but are prone to \textit{confirmation bias} as only a small portion of data is utilized in the early stages of training. Instead, we adaptively change the threshold according to the clients' learning status. In the early stages, we use a low threshold to include more data for training. As training progresses and the model becomes more confident, we increase the threshold to obtain more accurate pseudo-labels. We profile the learning status of each client and determine client-specific adaptive thresholds.

Recently, Sharpness-Aware Minimization (SAM) has demonstrated strong generalization capabilities across various tasks~\cite{abbas2022sharp, wang2023sharpness, liu2022towards}. Inspired by this, we hypothesized that applying SAM could effectively mitigate \textit{confirmation bias} among clients. However, our findings revealed that na\"ively applying SAM degrades performance. This issue occurs as SAM generalizes not only correctly pseudo-labeled samples, but also incorrectly pseudo-labeled ones. Generalization of incorrect data samples leads to the propagation of errors, thereby degrading the model's performance. Therefore, we apply consistency regularization to carefully selected data samples that are highly likely correct. We also uncover that the standard SAM objective (i.e., achieving flatter local minima) does not work well in FSSL. We thus propose a novel consistency regularization between the model outputs of adversarially perturbed and original weight parameters.

Finally, we propose a novel \textit{learning status-aware aggregation}. In FSSL, the learning difficulty can vary across clients. Since the server can access only a small labeled dataset, clients whose data closely resembles the server’s data will face lower learning difficulty. In comparison, those with more distinct data will encounter higher difficulty. Additionally, due to the non-iid data distribution of clients, the learning difficulty naturally differs among them. To account for different client learning difficulties, we assign higher aggregation weights to clients with higher learning difficulty, enabling the global model to learn more effectively from these clients. In contrast, previous FSSL approaches did not consider these variations in learning difficulty and relied on fixed aggregation weights.

Our main contributions are summarized as follows:
\begin{itemize}[topsep=3pt, align=left, labelwidth=10pt, leftmargin=15pt]
    \item We propose a \textit{client-specific adaptive threshold} that adjusts the pseudo-labeling threshold according to each client's learning status. By using a low threshold at the early stage of training, we effectively reduce confirmation bias by utilizing more data.
    \item We demonstrate that applying the SAM objective in FSSL is non-trivial and requires careful considerations. Minimizing the sharpness of incorrectly pseudo-labeled samples reduces the model performance. We also identify that the original SAM objective is ineffective in FSSL and propose a novel \textit{sharpness-aware consistency regularization} that regularizes consistency between original and perturbed model outputs.
    \item We propose \textit{learning status-aware aggregation} that adjusts the weight based on the client's learning status. Clients with lower learning status receive higher aggregation weights, ensuring their updates are well reflected in the global model.
    \item Our evaluation shows that our approach significantly outperforms existing methods across different settings, particularly when labeled data is scarce. \putname{} improves the classification accuracy of up to 23.0\% compared with existing FSSL methods.
\end{itemize}

\section{Related work}

\paragraph{Semi-supervised learning (SSL)} Recent SSL methods primarily stem from two key ideas: pseudo-labeling~\cite{pseudolb} and consistency regularization~\cite{consistencyregularization}. Pseudo-labeling artificially creates pseudo-labels and uses them as hard labels for supervised training. On the other hand, consistency regularization trains models by minimizing the variance between stochastic outputs, typically achieved through various weak or strong augmentations. FixMatch~\cite{fixmatch} generates high-quality pseudo labels via static probability thresholding and trains models to predict these labels from strongly augmented inputs. FlexMatch~\cite{flexmatch} enhances this approach by incorporating class-specific local thresholds alongside a fixed global threshold, adjusting based on the model's learning status. FreeMatch~\cite{wang2022freematch} dynamically adjusts the confidence threshold according to the model’s learning status and introduces a self-adaptive class fairness regularization penalty to encourage diverse predictions during early training. FlatMatch~\cite{huang2024flatmatch} increases generalization by adopting sharpness-awareness minimization~\cite{foret2020sharpness} into a cross-sharpness measure in SSL settings to ensure consistent learning performance between the labeled and unlabeled data.

\paragraph{Federated semi-supervised learning (FSSL)} Federated Learning (FL) enables collaborative training of a global model while ensuring data remains on the client side, thereby preserving data privacy (further discussed in Appendix~\ref{appendix:fl}). FSSL leverages FL in scenarios where labeled data is limited. FSSL research addresses two primary settings: \textit{labels-at-clients}~\cite{rscfed, fedssl, jeong2020federated, cbafed, zhang2024robust} and \textit{labels-at-server}~\cite{long2021fedcon, jeong2020federated, diao2022semifl}. In the \textit{labels-at-server} scenario, FedMatch~\cite{jeong2020federated} encourages similar outputs from similar clients using inter-client consistency loss. It employs disjoint training between the server and clients to mitigate forgetting issues. FedCon~\cite{long2021fedcon} utilizes contrastive learning to assist clients’ networks in learning embedding projections. SemiFL~\cite{diao2022semifl} achieves state-of-the-art results in the label-at-server setting by introducing \textit{alternate training}, which finetunes the global model with labeled data after each communication round. It generates pseudo-labels with the global model at the start of every communication round, rather than the common per-batch generation.

Real-world \textit{labels-at-server} FL scenarios to have extremely limited labeled data as labeling data requires domain expertise and could be costly~\cite{yang2021federated, elbir2022federated}. However, existing FSSL approaches target scenarios with hundreds of labeled data points (> 250) on the server, and their accuracy significantly deteriorates when only tens of labeled data points are available (Section~\ref{sec:perfcompfssl}). In contrast, \putname{} achieves high accuracy even in extremely label-scarce settings, such as when only 10 labeled data points are available on the server, demonstrating increased usability and practicality for real-world applications.
\section{Preliminaries}
\subsection{Federated learning}

Federated Learning~(FL) collaboratively trains a global model via coordinated communication with multiple clients. In communication round $t$, the server selects $K$ clients among available clients. The server transmits the current global model weights $W_g^t$ to selected clients. The selected clients update the model weight $W_k^t$ with the local dataset for $E$ epochs, where $k$ indicates the client index. Formally, $W^k_t = W^k_t - \eta \nabla_{W} \mathcal{L}_{\mathrm{client}}$, where $\mathcal{L}_{\mathrm{client}}$ denotes the objective function of clients, e.g., cross-entropy loss for the classification task. After local training, the server aggregates the trained model weights with $\beta^{k}$
 as aggregation weight of each client, which is
\begin{equation}
    \label{eq:fl_aggr}
    W^g_{t+1} = \sum_{k=1}^{K} \beta^{k} W^k_t.
\end{equation}

\subsection{Federated semi-supervised learning}

In Federated Semi-Supervised Learning~(FSSL), especially in the \textit{labels-at-server} scenario, 
labeled dataset $\mathcal{D}^S_L = \{(x_{b}, y_{b}) : b \in [N_{L}]\}$ is only available at the server, while clients have only unlabeled dataset $\mathcal{D}^k_U = \{u_{b} : b \in [N^{k}_U]\}$, where $N_{L}$ and $N_{U}=\sum\nolimits_{k=1}^{K} N^{k}_{U}$ are the total number of labeled data and unlabeled data, respectively. In general, $N_{L} \ll N_{U}$. At each communication round $t$, the server updates its model weight $W^S_t$ with supervised loss $\mathcal{L}_\mathrm{server}$ for $E$ local epochs with
\begin{equation}
    \label{eq:fssl_server_train}
    \mathcal{L}_{\mathrm{server}} = \frac{1}{B} \sum^{B}_{b=1} \mathcal{H}(y_{b}, p_{W^S_t}(y|w(x_{b}))),
     \quad W^S_t = W^S_t - \eta \nabla_{W} \mathcal{L}_{\mathrm{server}},
\end{equation}
where data batch $(x_b, y_b)$ is randomly drawn from $\mathcal{D}^S_L$ with batch size $B$. $\mathcal{H}(\cdot , \cdot)$ refers to the cross-entropy loss, $\omega(\cdot)$ is the weak data augmentation (e.g., random horizontal flip and crop), and $p_W(\cdot)$ is the output probability from model $W$. Clients update their model weight $W^k_t$ using cross-entropy loss with pseudo-labeling, which can be expressed as
\begin{equation}
    \mathcal{L}_\mathrm{client} = \frac{1}{\mu B} \sum^{\mu B}_{b=1} \mathbbm{1} (\max(q_b) > \tau) \cdot \mathcal{H}(\hat{q_b}, Q_b),
     \quad W^k_t = W^k_t - \eta \nabla_{W} \mathcal{L}_{\mathrm{client}},
\end{equation}
where $q_b$ and $Q_b$ are the abbreviations of $p_{W^k_t}(y|\omega(u_b))$ and $p_{W^k_t}(y|\Omega(u_b))$, respectively. Data batch $u_b$ is randomly selected from $\mathcal{D}^k_U$ with a batch size of $B$. The one-hot label form of $q_b$ is denoted as $\hat{q}_b$, and the ratio of data with confidence above $\tau$ is represented by $\mu$. The indicator function $\mathbbm{1}(\cdot > \tau)$ is used for confidence-based thresholding. $\Omega(\cdot)$ represents strong augmentation (e.g., RandAugment \cite{cubuk2020randaugment}).

We adopt ``fine-tune global model with labeled data'' and ``generate pseudo-labels with global model'' strategies from SemiFL~\cite{diao2022semifl}. In communication round $t$, the server distributes the current global model $W_t^g$ to $K$ selected clients. Before training, clients generate pseudo-labels for a local dataset with a fixed global model $W_t^g$. The changed local objective function is 
\begin{equation}
    \mathcal{L}_\mathrm{client} = \frac{1}{\mu B} \sum^{\mu B}_{b=1} \mathbbm{1} (\max(q^g_b) > \tau) \cdot \mathcal{H}(\hat{q}^g_b, Q_b),
\end{equation}
where $q_b^g$ stands for $p_{W_t^g}(y|\omega(u_b))$. Subsequently, the server aggregates trained local models with Eq~\ref{eq:fl_aggr}. The server fine-tunes the aggregated model with $\mathcal{L}_\mathrm{server}$, yielding a new global model $W^g_{t+1}$.

\subsection{Sharpness-aware minimization}

Sharpness-Aware Minimization~(SAM)~\cite{foret2020sharpness, kwon2021asam} has been increasingly applied to various tasks~\cite{abbas2022sharp, wang2023sharpness, liu2022towards} due to its ability to enhance generalization. SAM improves generalization by minimizing the sharpness of the loss landscape, which helps in finding flatter minima that generalize better across different tasks and datasets. Traditional optimization methods could lead to sharp minima, resulting in poor generalization to unseen data. SAM addresses this issue by incorporating weight perturbations into the optimization objective to find flatter minima. 
The core objective of SAM is defined as:
\begin{equation}
    \min_{w} \max_{\|\epsilon\|_2 < \rho} \mathcal{L}_{w + \epsilon},
\end{equation}
where \( \epsilon \) is a perturbation vector constrained within a \(\rho\)-ball around the model weight \( w \). The inner maximization seeks to find the perturbation \( \epsilon \) that maximizes the loss \(\mathcal{L}\) within the specified \(\rho\)-ball.

To make this optimization feasible, SAM approximates the perturbation \( \epsilon \) as:
\begin{equation}
    \epsilon^* = \rho \frac{\nabla_w \mathcal{L}_w}{\|\nabla_w \mathcal{L}_w\|_2}.
\end{equation}

This approximation simplifies the inner maximization by scaling the gradient direction to have a norm of \(\rho\). The outer minimization updates the weights using the gradient evaluated at the perturbed weights \( w + \epsilon^* \). Specifically, the gradient used for the weight update is $\nabla_w \mathcal{L}_{w + \epsilon^*}$.

\begin{figure}[t]
    \centering
    \includegraphics[width=1.0\textwidth]{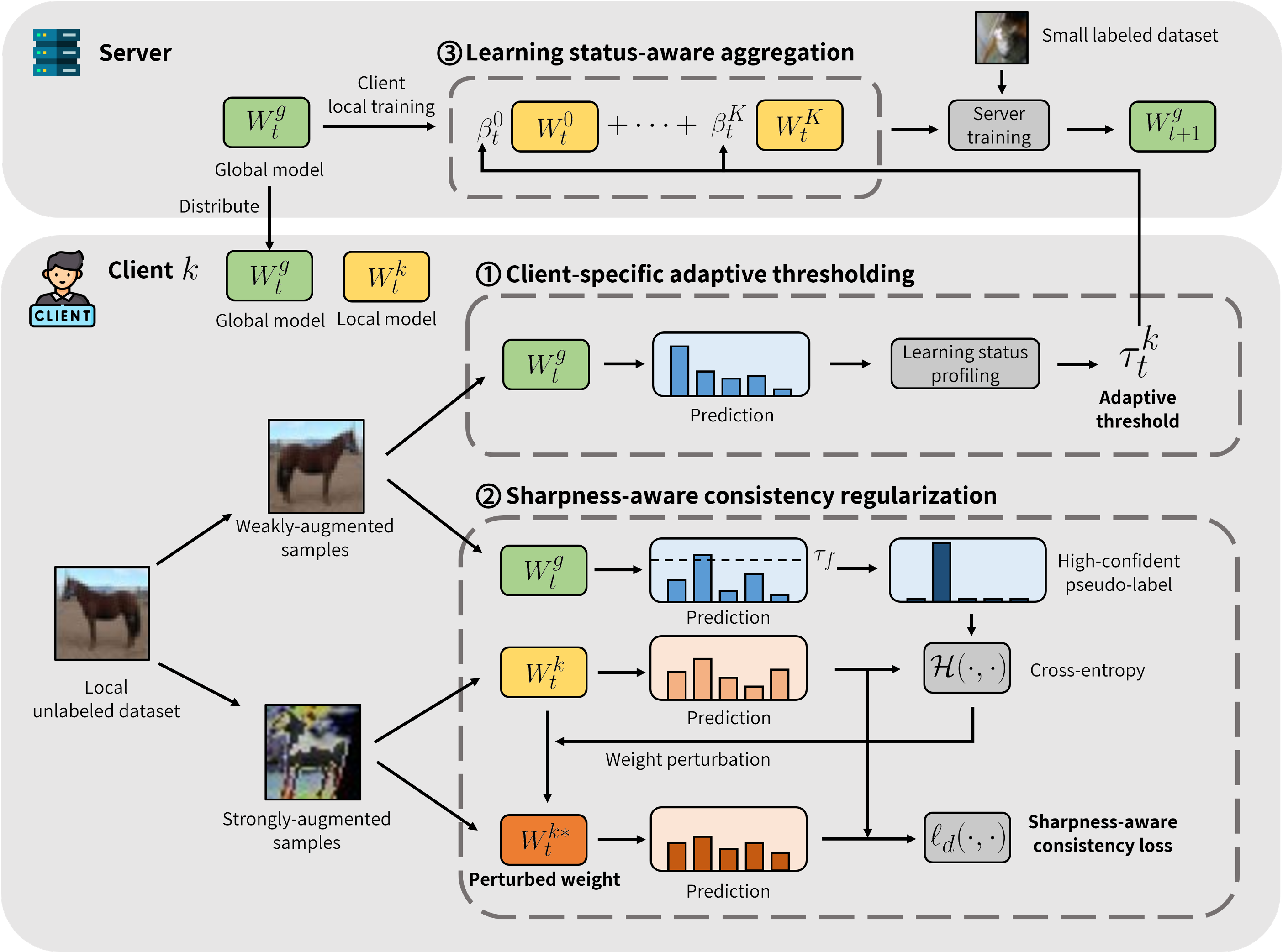}
    \caption{Overview of \putname{}:
(1) \textit{client-specific adaptive thresholding} adjusts the pseudo-labeling threshold according to each client's learning status,
(2) \textit{sharpness-aware consistency regularization} ensures consistency between the original model and the adversarially perturbed model with carefully selected high-confident pseudo labels, and
(3) \textit{learning status-aware aggregation} aggregates client models considering each client's learning progress.}
    
    \label{fig:overview}
\end{figure}

\section{Method}

\textbf{F}ew-\textbf{L}abels \textbf{F}ederated semi-supervised \textbf{L}earning, abbreviated as \putname{}, has three key components: (1)~\textit{client-specific adaptive thresholding}, which leverages more unlabeled data by dynamically adjusting thresholds for pseudo-labeling, (2)~\textit{sharpness-aware consistency regularization}, which minimizes sharpness for carefully selected data to ensure better generalization, and (3)~\textit{learning status-aware aggregation}, which aggregates local models from clients while considering their learning progress. Fig.~\ref{fig:overview} overviews \putname{} and Appendix~\ref{appendix:algorithm} details the algorithm.

\subsection{Client-specific adaptive thresholding}
\label{section:cat}

We use an adaptive thresholding mechanism rather than a fixed threshold to incorporate more unlabeled data from the beginning of training. This approach is inspired by FreeMatch~\cite{wang2022freematch} that gradually increases the threshold according to the model learning status. At round $t$, each client profiles its learning status during the pseudo-label generation stage using local unlabeled dataset $\mathcal{D}^k_U$ with global model $W_t^g$. Adaptive threshold $\tau^k_t$ of client $k$ at round $t$ is
\begin{equation}
    \tau_t^k = \frac{1}{|\mathcal{D}^k_U|} \sum^{|\mathcal{D}^k_U|}_{b=1}\max(q^g_b),
\end{equation}
where $q^g_b$ is $q_b$ calculated with global model $W_t^g$. This approach sets a low initial threshold value, as the model exhibits lower confidence in the data at the beginning of training. The threshold gradually increases as training progresses, allowing the model to focus on high-confidence data. Additionally, we estimate the learning status specific to each class and apply different thresholds for each class. This is achieved by utilizing the output probabilities of the global model's predictions for each class, which can be expressed as
\begin{equation}
    \label{eq:pk}
    {\tilde{p}}_{t}^{k}(c) = \frac{1}{|\mathcal{D}^k_U|} \sum^{|\mathcal{D}^k_U|}_{b=1} q_b^g(c).
\end{equation}
We calculate client-specific adaptive thresholds for each class using $\tau_{t}^{k}$ and ${\tilde{p}}_{t}^{k}(c)$ as
\begin{equation}
   \tau_{t}^{k}(c) = \mathrm{MaxNorm}( {\tilde{p}}_{t}^{k}(c)) \cdot \tau_t^k = \frac{ {\tilde{p}}_{t}^{k}(c)}{\max\{ {\tilde{p}}_{t}^{k}(c) : c \in [C]\}} \cdot \tau_{t}^{k}.
\end{equation}
The unsupervised training objective $\mathcal{L}_a$ of client $k$ with adaptive thresholding at each iteration is:
\begin{equation}
    \mathcal{L}_a^k = \frac{1}{\mu B} \sum^{\mu B}_{b=1} \mathbbm{1} (\max(q^g_b) > \tau_{t}^{k}(\mathrm{arg max}(q^g_b)) \cdot \mathcal{H}(\hat{q}^g_b, Q_b).
\end{equation}

\subsection{Sharpness-aware consistency regularization}
\label{section:scr}

While Sharpness-Aware Minimization~(SAM) generalizes well in many tasks~\cite{abbas2022sharp, wang2023sharpness, liu2022towards}, it is not trivial to apply it to FSSL, as SAM generalizes not only correctly pseudo-labeled samples but also incorrect samples. This indiscriminate generalization results in the propagation of errors, thereby degrading the model's performance~(Section~\ref{eval:wrong}). To tackle this issue, we apply consistency regularization to a carefully curated subset of data samples with a high confidence of correctness. While we use client-specific adaptive threshold~(Section~\ref{section:cat}), we use a high fixed threshold to get high-confidence data samples. \putname{} adversarially perturbs the weight parameters that maximize loss calculated with high-confidence data samples and regularizes consistency using the perturbed weight. 

\textbf{Adversarial weight perturbation}\quad When a client $k$ trains its local model $W^k$ with mini-batch, the model weight is perturbed with $\epsilon^*$ that increases $\mathcal{L}_p^k$ the most, where $\epsilon^*$ and $\mathcal{L}_p^k$ are defined as
\begin{equation}
    \mathcal{L}_p^k = \frac{1}{\mu B} \sum^{\mu B}_{b=1} \mathbbm{1} (\max(q_b^g) > \tau_f) \cdot \mathcal{H}(\hat{q}_b^g, Q_b),
\end{equation}
\begin{equation}
    \label{eq:perturb}
    \epsilon_p^* = \argmax_{\|\epsilon\|_2 \leq \rho} \mathcal{L}_p^k \approx \rho \frac{\nabla_{W^k} \mathcal{L}_p^k}{\|\nabla_{W^k} \mathcal{L}_p^k\|_2},
    \quad W^{k*} = W^k + \epsilon_p^*.
\end{equation}
where $\rho$ stands for perturbation strength. We use a large fixed threshold $\tau_f$ to get a high-confidence pseudo-label.

\textbf{Consistency regulation}\quad With the perturbed weight $W^{k*}$, we calculate $Q_b^*$, which is the output probability of a strongly augmented sample for $W^{k*}$. Unlike traditional SAM objective that takes $\nabla_{W^{k*}}\mathcal{L}_p$, we measure the difference of model outputs between the original and the perturbed models~(Section~\ref{eval:sam}). Formally, 
\begin{equation}
    \mathcal{L}_{cs}^k= \ell_d(Q_b^*, Q_b), \textrm{where } Q_b^*=p_{W^{k*}}(y|\Omega(u_b)),
\end{equation}
in which $\ell_d(\cdot , \cdot)$ measures the distance (e.g., L2 distance or KL divergence). Finally, local training objective of client $k$ with client-specific adaptive thresholding~(Section~\ref{section:cat}) and sharpness-aware consistency regularization is

\begin{equation}
    \mathcal{L}_\mathrm{client}^k = w_a \mathcal{L}_a^k + w_{cs} \mathcal{L}_{cs}^k 
\end{equation}
with $w_a$ and $w_{cs}$ being the loss weights. \putname{} effectively leverages both low-confidence data with \textit{client-specific adaptive threshold} and high-confidence data with \textit{sharpness-aware consistency regularization} to minimize the confirmation bias of clients.

\subsection{Learning status-aware aggregation}

After the local training of the selected $K$ clients, the server aggregates the trained local models using weights $\beta^k$, as shown in Eq. \ref{eq:fl_aggr}. While existing FSSL approaches use uniform weights $(\beta^k=1/K)$, we propose a \textit{learning status-aware aggregation} that adjusts the aggregation weight based on the client's learning status. For a client with a low learning status, indicated by a low adaptive threshold $\tau_t^k$, we increase the aggregation weight so that the local learning is better reflected in the global model. We calculate the aggregation weight as
\begin{equation}
    \beta_t^k = \frac{1-\tau_t^k}{\sum_{k=1}^K(1-\tau_t^k)}.
\end{equation}

Our aggregation method complements the client-specific adaptive thresholds~(Section \ref{section:cat}). In this scheme, we use lower thresholds for clients with a lower learning status to enable more extensive learning from their data. By extending this notion to the client level, clients with lower thresholds, which indicate more valuable learning updates, are given a greater influence on the global model. This ensures that the most informative updates are prioritized.

\section{Experiments}

\subsection{Setup}
\label{exp:setup}

\paragraph{Data setup} We evaluate \putname{} in three public datasets: CIFAR10, CIFAR100~\cite{alex2009learningcifar}, and SVHN~\cite{svhn}. We test our method under balanced IID and unbalanced non-IID data distribution settings. Each client receives an equal amount of unlabeled data in the balanced IID setting. We sample data using a Dirichlet distribution $\mathrm{Dir}(\alpha)$ for the unbalanced non-IID setting. Each client receives a different number of data samples and samples per class. As $\alpha \rightarrow \infty$, the distribution approaches IID. We set $\alpha=\{0.1,0.3\}$ in our experiments. The number of labeled data samples at the server~($N_L$) is set to $\{10, 40\}$ for CIFAR10, $\{100, 400\}$ for CIFAR100, and $\{40, 250\}$ for SVHN, following widely-used evaluation settings for SSL~\cite{wang2022freematch, huang2024flatmatch}.

\paragraph{Learning setup} In our experiments, we use 100 clients, with a participation ratio of 0.1 per communication round~($K=10$). We adopt the WideResNet~\cite{zagoruyko2016wide} as our backbone, employing WideResNet28x2 for the CIFAR10 and SVHN datasets, and WideResNet28x8 for the CIFAR100 dataset. Both the server and clients optimize their local datasets for five local epochs, with 800 communication rounds. We employ the momentum SGD optimizer with a learning rate of 0.03, momentum of 0.9, and weight decay of 5e-4, following previous work~\cite{diao2022semifl}. For sharpness-aware consistency regularization (Section~\ref{section:scr}), we use the KL-divergence loss function for $\ell_d(\cdot , \cdot)$. For adversarial weight perturbation (Eq. \ref{eq:perturb}), we use ASAM~\cite{kwon2021asam}, which implements scale invariance on standard SAM~\cite{foret2020sharpness}. Based on a grid search, the perturbation strength $\rho$ is set to 0.1 for the CIFAR10 and SVHN datasets and 1.0 for the CIFAR100 dataset. For strong data augmentation, we use RandAugment~\cite{cubuk2020randaugment}. We also adopt the static Batch Normalization (sBN)~\cite{diao2020heterofl} strategy, as utilized in SemiFL. Further details on sBN are in Appendix \ref{appendix:sbn}. We used RTX3090 GPUs throughout the experiment. Additional details are in Appendix~\ref{appendix:learning}. 

\renewcommand{\arraystretch}{1.2}
\begin{table}[ht]
    \captionsetup{skip=5pt}
    \centering
    \caption{Evaluation of \putname{} compared with existing FSSL methods. We report the average accuracy(\%) and standard deviation across three runs with different random seeds. \putname{} shows significant performance improvements over existing methods across different settings. \textbf{Bold} indicates the best result and \underline{underline} indicates the second-best result.}
    \label{tab:result}
    \begin{adjustbox}{max width=0.9\textwidth}
        \begin{tabular}{cccccccc}
        \toprule
        \multicolumn{2}{c}{Dataset}                                                                     & \multicolumn{2}{c}{CIFAR10}    & \multicolumn{2}{c}{SVHN}     & \multicolumn{2}{c}{CIFAR100}                                        \\ 
        \cmidrule(lr){3-4} \cmidrule(lr){5-6} \cmidrule(lr){7-8}
\multicolumn{2}{c}{\# of labeled data samples ($N_L$)}                                                  & 10             & 40            & 40             & 250         & 100              & 400\\
        \midrule
        \multirow{4}{*}{\makecell{Unbalanced Non-IID,\\ Dir(0.1)}} 
                                 & FedMatch                                   & 16.0(2.3)     & \underline{25.6(2.2)}     & \underline{20.7(2.7)}     & 70.1(2.2)   & 6.3(0.3) & 10.0(1.8) \\
                                                      & FedCon                    & \underline{16.6(2.1)}    & 25.4(2.3)    & 20.5(1.4)    & 73.1(2.0) & 4.0(0.4) & 8.2(0.6)   \\
                                                      & SemiFL                    & 10.0(0.0)     & 19.9(7.5)     & 18.0(2.6)      & \underline{82.3(1.8)}    & \underline{9.8(2.4)} & \underline{13.5(5.0)}\\
                                           \rowcolor{LightCyan}
                                                      & \putname{}                                      & \textbf{19.2(5.7)}    & \textbf{36.4(1.4)}   & \textbf{21.5(3.3)}     & \textbf{88.0(1.0)}   & \textbf{10.4(1.3)} & \textbf{23.5(1.2)}\\
        \midrule
        \multirow{4}{*}{\makecell{Unbalanced Non-IID,\\ Dir(0.3)}} 
                                 & FedMatch                                   & 15.3(1.3)     & 25.2(3.5)     & 22.3(0.7)     & \underline{72.3(3.0)}   & 5.5(1.5) & 9.8(1.1) \\
                                                      & FedCon                    & \underline{16.9(2.4)}    & 26.5(2.1)     & 21.6(1.7)     & 68.7(2.7)   & 5.8(0.6) & 13.3(0.9) \\
                                                      & SemiFL                   & 10.0(0.0)     & \underline{38.0(2.7)}     & \underline{26.3(2.5)}      & 42.7(40.1)    & \textbf{12.4(1.2)} & \underline{18.9(9.7)}\\
                                           \rowcolor{LightCyan}
                                                          & \putname{}                                  & \textbf{24.3(4.5)}    & \textbf{43.5(7.5)}   & \textbf{31.0(4.2)}     & \textbf{92.6(0.5)}   & \underline{12.1(1.1)} & \textbf{25.4(1.0)}\\
        \midrule
        \multirow{4}{*}{Balanced IID}              
                                & FedMatch                                    & 16.2(1.9)    & 25.4(2.8)    & 18.4(4.7)   & 66.2(0.8)   & 6.4(0.6) & 10.0(1.7)\\
                                                      & FedCon                   & \underline{16.7(2.0)}  & 23.3(6.2)   & 20.3(1.0)  & \underline{71.6(1.5)} & 5.7(0.6) & 12.4(1.6)\\
                                                       & SemiFL                   & 10.0(0.0)   & \underline{75.3(2.8)}   & \underline{53.4(13.3)}     & 43.3(41.0)   & \underline{13.9(3.3)} & \underline{27.9(6.1)}\\
                                           \rowcolor{LightCyan}
                                                      & \putname{}                                      & \textbf{38.9(11.1)}    & \textbf{81.5(7.4)}   & \textbf{75.3(2.4)}   & \textbf{94.6(1.1)}   & \textbf{14.4(2.3)} & \textbf{28.1(2.2)}\\  
        \bottomrule
    
\end{tabular}
\end{adjustbox}
\end{table}

\subsection{Performance comparison with FSSL algorithms}
\label{sec:perfcompfssl}

We evaluate \putname{} against existing FSSL methods: FedMatch~\cite{jeong2020federated}, FedCon~\cite{long2021fedcon}, and SemiFL~\cite{diao2022semifl}. Table~\ref{tab:result} shows that \putname{} consistently delivers the best or nearly the best performance 
across all settings. 
For instance, although SemiFL performs the best in the non-IID-0.3 setting of CIFAR100 with 100 labels, it struggles to generalize to other scenarios. SemiFL achieves only around 10\% accuracy in CIFAR10 with 10 labels and about 43\% accuracy in SVHN with 250 labels. In contrast, \putname{} consistently maintains high accuracy across all tasks. The performance gap compared with the best-performing algorithm (SemiFL) in non-IID-0.3/CIFAR100/100-labels is only 0.3\%. Except for that, \putname{} consistently outperforms the baseline methods across all other settings. Additionally, \putname{} demonstrates a substantial improvement over existing methods, achieving \textbf{20.3\%} higher performance in non-IID-0.3/SVHN/250-labels and \textbf{23.0\%} higher performance in IID/SVHN/250-labels. These findings indicate that \putname{} effectively mitigates \textit{confirmation bias} among clients, leading to robust generalization even with limited data across different settings.

We emphasize that \putname{} significantly outperforms other methods when labeled data is extremely limited: by \textbf{22.2\%} on the IID setting of CIFAR10 with 10 labels and by \textbf{21.9\%} on the IID setting of SVHN with 40 labels. This substantial margin highlights \putname{}'s exceptional ability to leverage scarce labeled data, making it practical for real-world federated learning applications. Additional experiments are provided in Appendix~\ref{appendix:more-exp}.

\begin{table}[h]
    \captionsetup{skip=5pt}
    \centering
    \caption{Contribution of each component of \putname{} on the SVHN dataset ($N_L=40$, balanced IID). By applying Client-specific Adaptive Thresholding (CAT) and Sharpness-Aware Consistency Regularization (SACR) to the baseline (FixMatch + FedAvg), performance is boosted. The combination of CAT and SACR further improves the accuracy. Incorporating Learning Status-Aware Aggregation (LSAA) leads to the best performance, finally achieving \putname{}. The result demonstrates the importance of each component in \putname{}.}
    \label{table:component}
    \begin{adjustbox}{max width=0.4\textwidth}
    \begin{tabular}{lc}
    \toprule
    \multicolumn{1}{c}{Algorithm} & Accuracy \\
    \midrule
    FixMatch + FedAvg             & 50.2     \\
    \midrule
    \textbf{SACR} + FixMatch + FedAvg               & 60.9     \\
    \textbf{CAT} + FedAvg                  & 68.2     \\
    \textbf{CAT} + \textbf{SACR} + FedAvg           & 71.7     \\
    \rowcolor{LightCyan}
    \putname{}: \textbf{CAT} + \textbf{SACR} + \textbf{LSAA}             & \textbf{73.2}     \\
    \bottomrule
    \end{tabular}
    \end{adjustbox}
\end{table}
\vspace{-1em}

\paragraph{Significance of each component of \putname{}} We assess the contribution of each component of \putname{}: Client-specific Adaptive Thresholding (CAT), Sharpness-Aware Consistency Regularization (SACR), and Learning Status-Aware Aggregation (LSAA) in Table~\ref{table:component}. The accuracy improvements provided by each component are evaluated using the SVHN dataset with 40 labeled data points and a balanced IID setting. We use FixMatch + FedAvg as the baseline, where FixMatch~\cite{fixmatch} employs a fixed threshold for pseudo-labeling. Our results indicate that both SACR and CAT significantly enhance the performance over the baseline. Combining SACR and CAT yields further accuracy improvements. Finally, integrating LSAA for model aggregation, equivalent to \putname{}, achieves the highest accuracy. These findings demonstrate that each component of \putname{} contributes uniquely and complementarily to the overall performance.

\begin{figure}[h]
    \centering
    \begin{subfigure}{0.32\textwidth}
        \includegraphics[width=\textwidth]{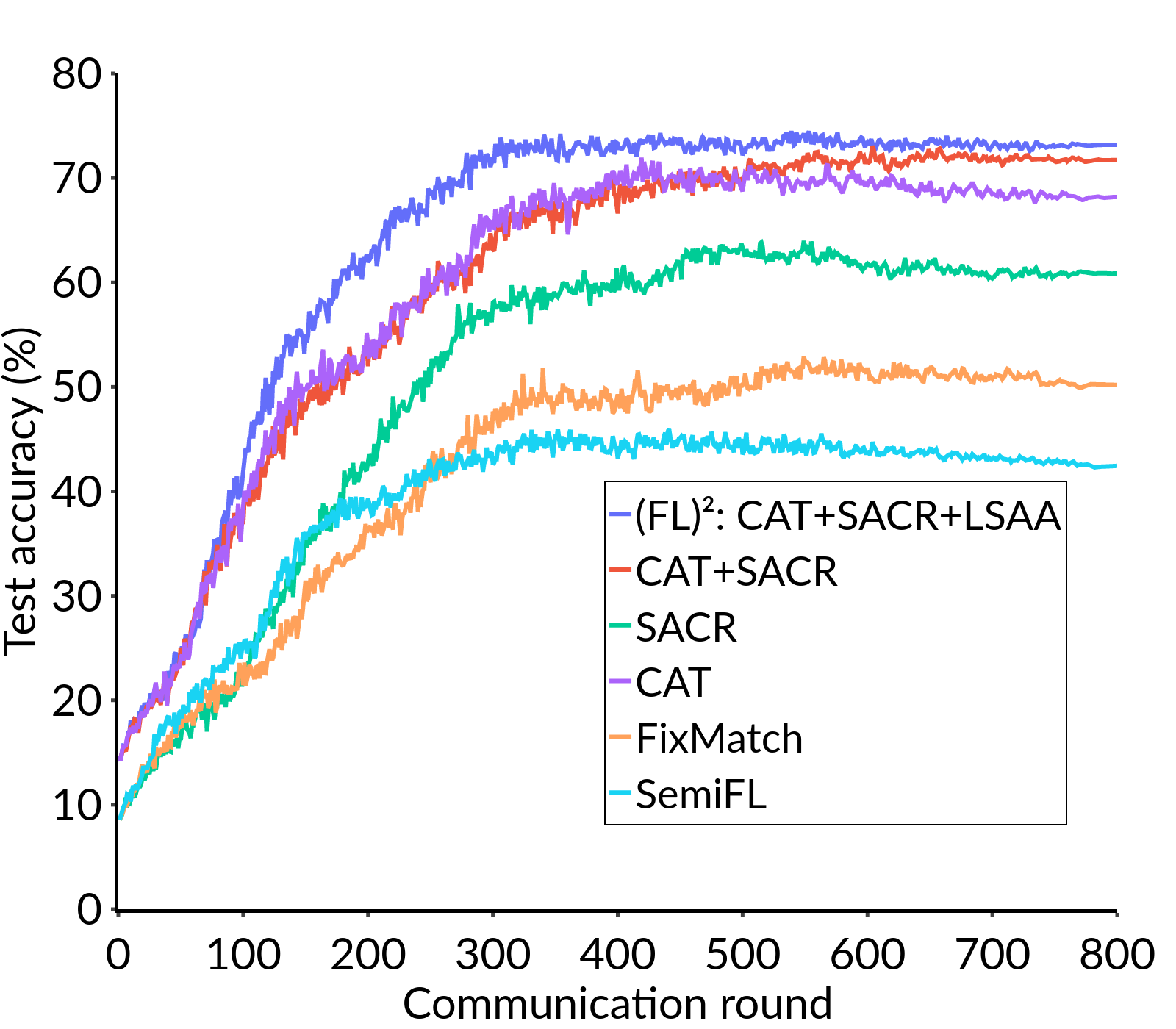}
        \caption{Test accuracy.}
        \label{fig:cb-testacc}
    \end{subfigure}
    \begin{subfigure}{0.32\textwidth}
        \includegraphics[width=\textwidth]{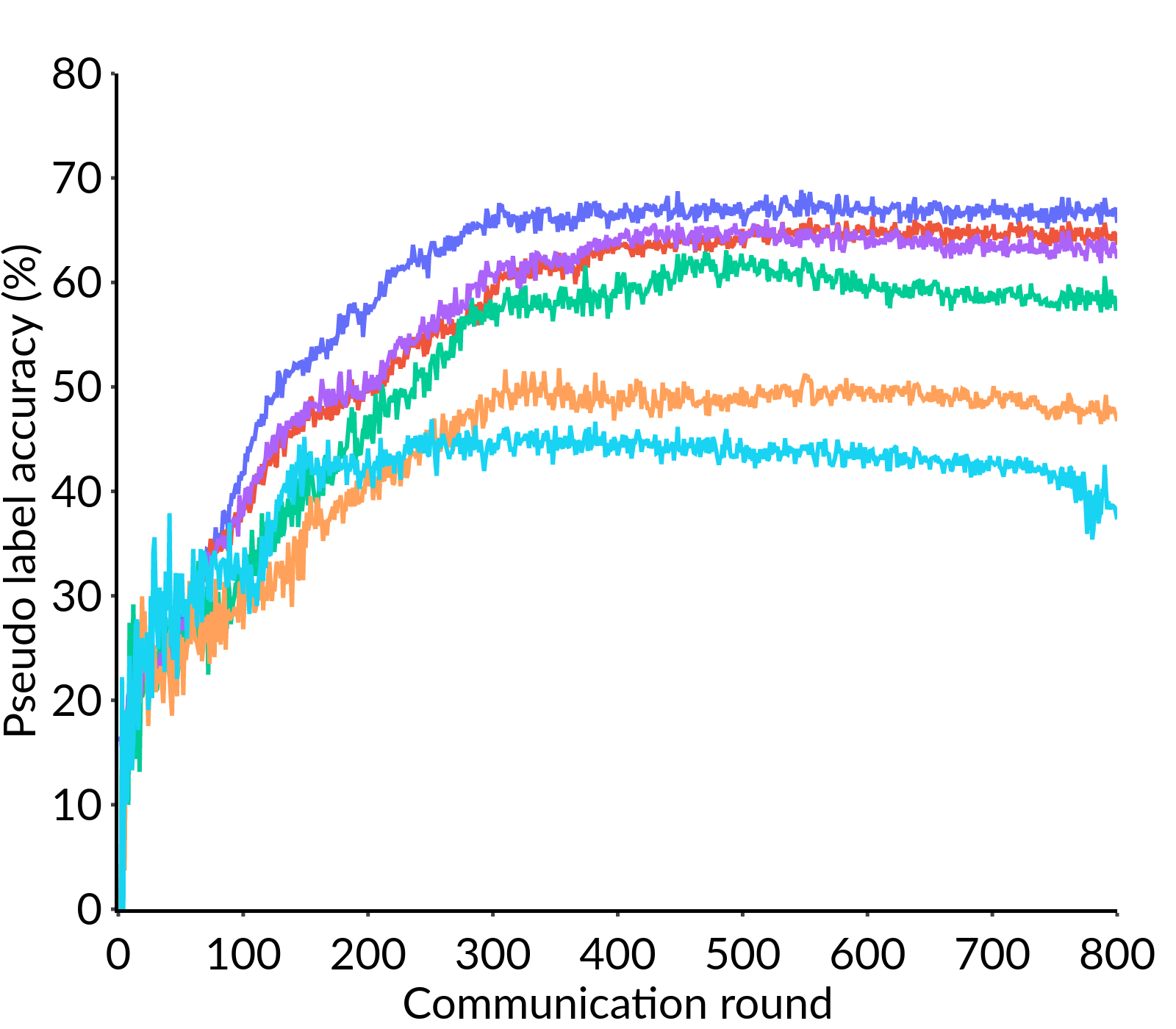}
        \caption{Pseudo label accuracy.}
        \label{fig:cb-pseudoacc}
    \end{subfigure}
    \begin{subfigure}{0.32\textwidth}
        \includegraphics[width=\textwidth]{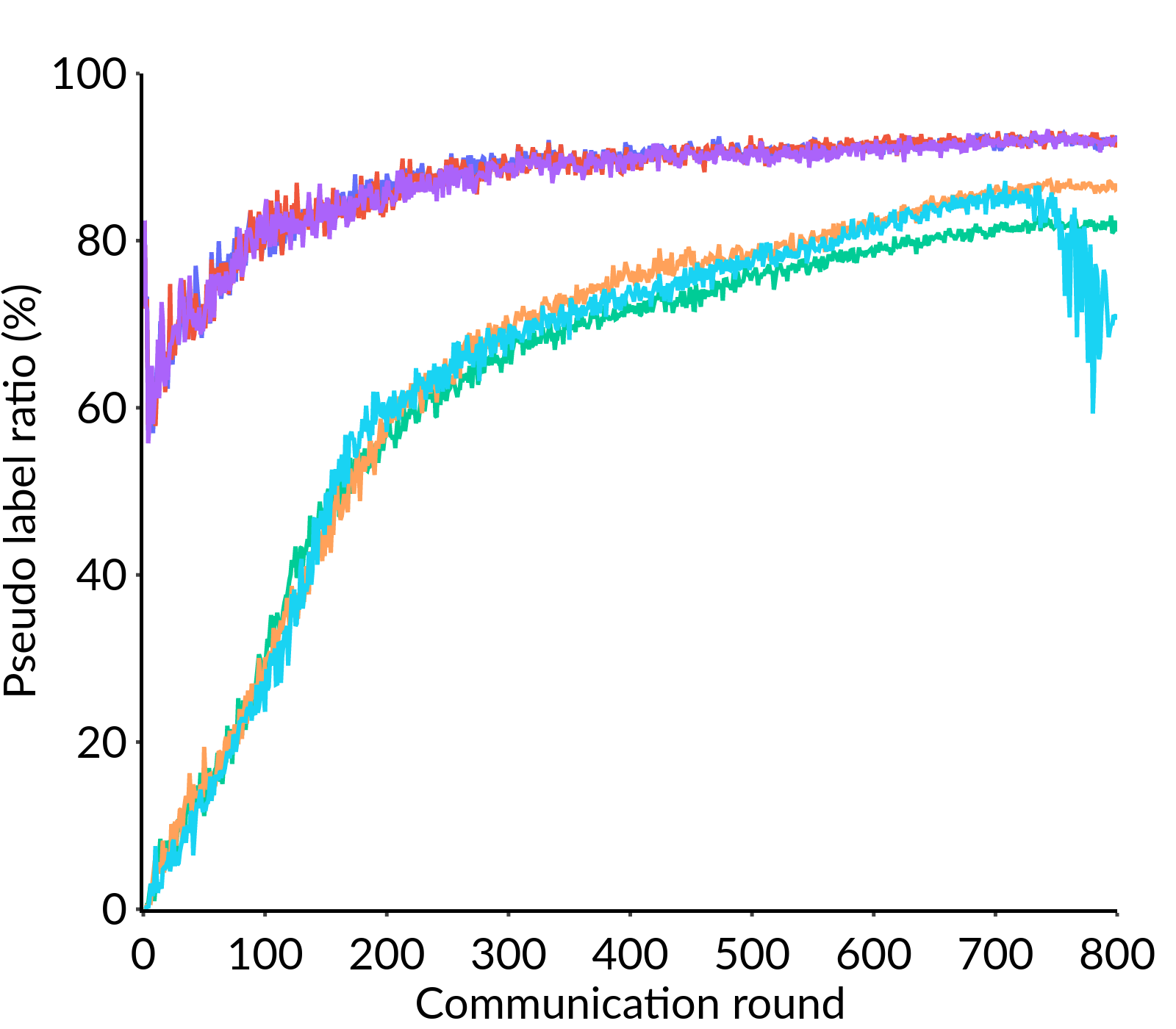}
        \caption{Pseudo label ratio.}
        \label{fig:cb-pseudoratio}
    \end{subfigure}\\ 
    \begin{subfigure}{0.32\textwidth}
        \includegraphics[width=\textwidth]{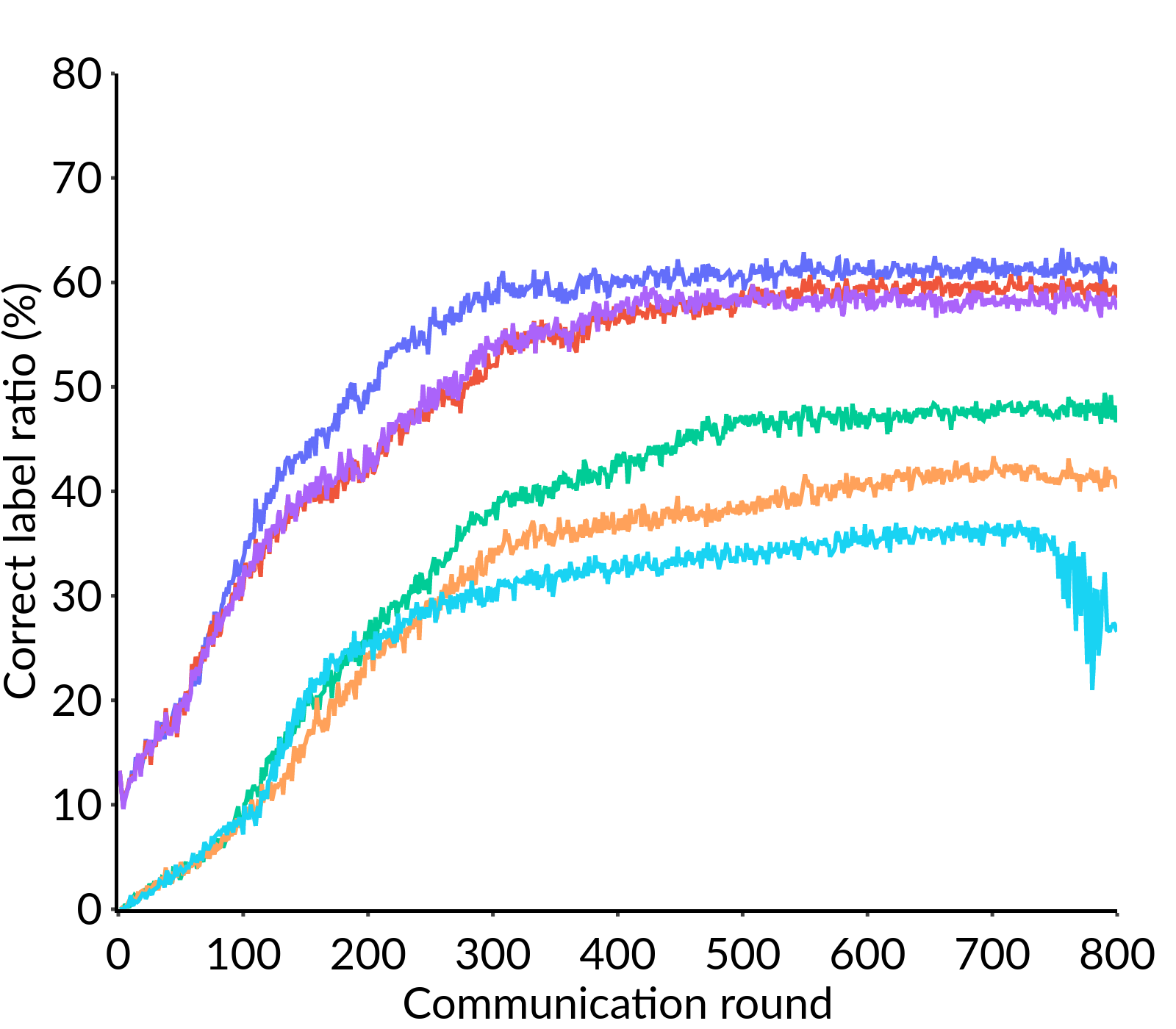}
        \caption{Correct label ratio.}
        \label{fig:cb-correctratio}
    \end{subfigure}
    \begin{subfigure}{0.32\textwidth}
        \includegraphics[width=\textwidth]{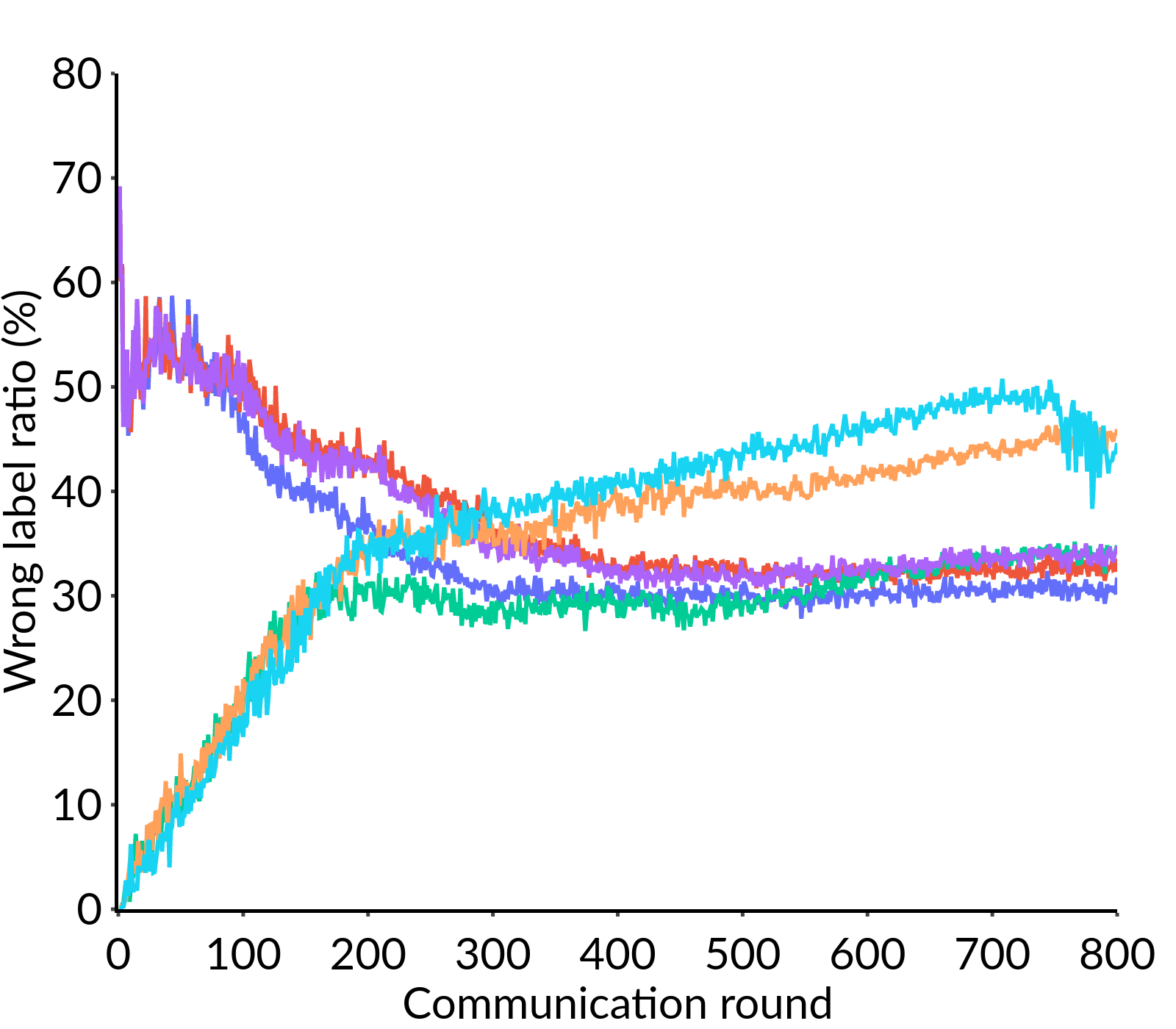}
        \caption{Wrong label ratio.}
        \label{fig:cb-wrongratio}
    \end{subfigure}
    \begin{subfigure}{0.32\textwidth}
        \includegraphics[width=\textwidth]{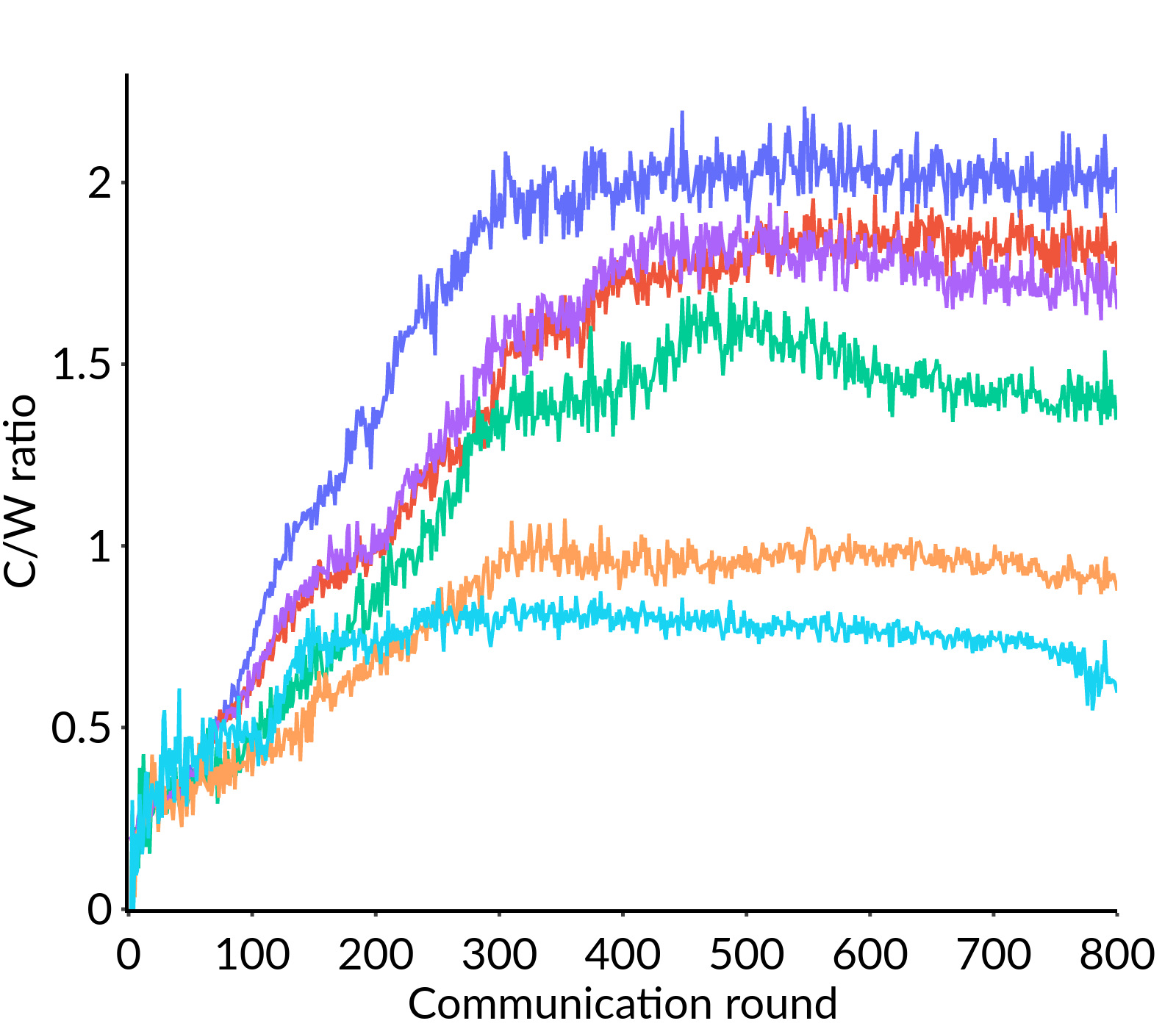}
        \caption{C/W ratio.}
        \label{fig:cb-cwratio}
    \end{subfigure}
    
    \caption{Comparison of SemiFL, \putname{}, and its variants on the SVHN dataset ($N_L = 40$, balanced IID). Pseudo-label accuracy measures the percentage of correct pseudo-labels. The label ratio is the proportion of pseudo-labeled samples among all unlabeled data. Correct and wrong label ratios indicate the percentages of correctly and incorrectly labeled samples, respectively. The C/W ratio shows the number of correct labels relative to wrong labels. All subgraphs share the legend of Fig.~\ref{fig:cb-testacc}.}
    \label{fig:cb}
    \vspace{-1em}
\end{figure}

\subsection{Effect of \putname{} on confirmation bias}
Since incorrect pseudo-labels usually lead to confirmation bias~\cite{arazo2020pseudo}, we evaluated pseudo-label accuracy, label ratio, correct label ratio, wrong label ratio, and C/W ratio in addition to test accuracy. We compared \putname{} against baseline methods using the SVHN dataset with 40 labels in a balanced IID setting, as reported in Fig.~\ref{fig:cb}.
A high pseudo-label accuracy indicates that the method produces reliable pseudo-labels. A high correct label ratio suggests that the method supplies the model with more accurate labels. Conversely, a low wrong label ratio indicates that the model encounters fewer incorrect labels, which is crucial for minimizing confirmation bias~\cite{arazo2020pseudo}. Lastly, a high C/W ratio signifies that the model is exposed to more correct labels than incorrect ones, further helping to reduce confirmation bias.

We observed that \putname{} consistently outperforms SemiFL across all metrics. While SemiFL generates more incorrect labels (C/W ratio < 1), \putname{} produces twice as many correct labels than incorrect ones~(Fig.~\ref{fig:cb-cwratio}). Additionally, the wrong label ratio for \putname{} is approximately 30\%, significantly lower than SemiFL’s 45\%~(Fig.~\ref{fig:cb-wrongratio}). These results suggest that \putname{} effectively reduces incorrect pseudo-labels while increasing correct ones, thereby mitigating confirmation bias. Furthermore, we observe the effectiveness of each component of \putname{}, which are CAT, SACR, and LSAA. Using CAT and SACR alone delivers better performance than the baseline for all metrics. If we use CAT + SACR, pseudo label accuracy increases, correct label ratio increases, and wrong label ratio decreases, which means we reduce the confirmation bias. When LSAA is added, which is \putname{}, it achieves the best performance across all metrics. This suggests that the synergistic effect of CAT, SACR, and LSAA effectively reduces confirmation bias.

\begin{figure}[h]
    \centering
    \begin{subfigure}{0.35\textwidth}
        \includegraphics[width=\textwidth]{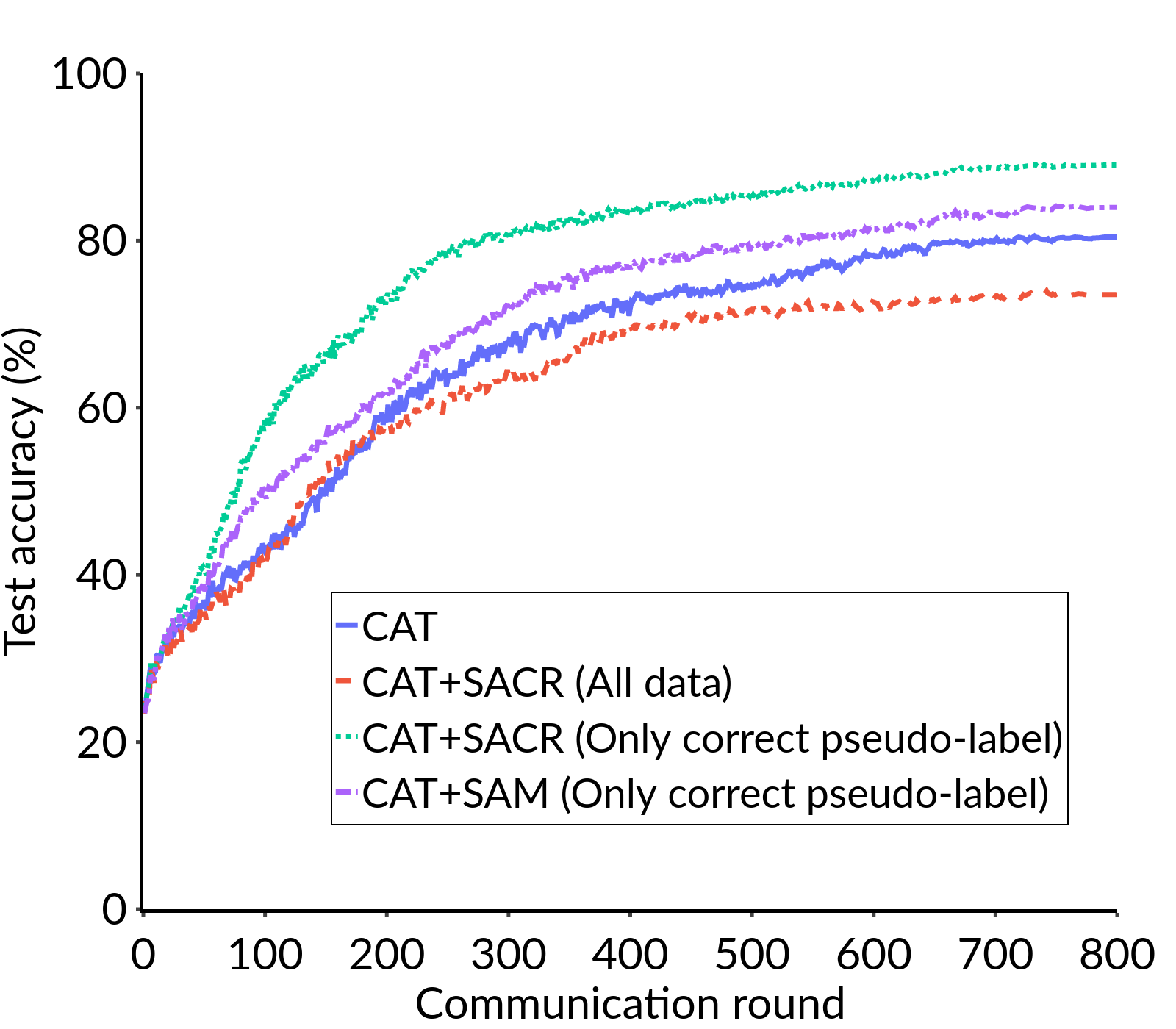}
        \caption{Test accuracy.}
        \label{fig:wrong-pseudo-1}
    \end{subfigure}
    \begin{subfigure}{0.35\textwidth}
        \includegraphics[width=\textwidth]{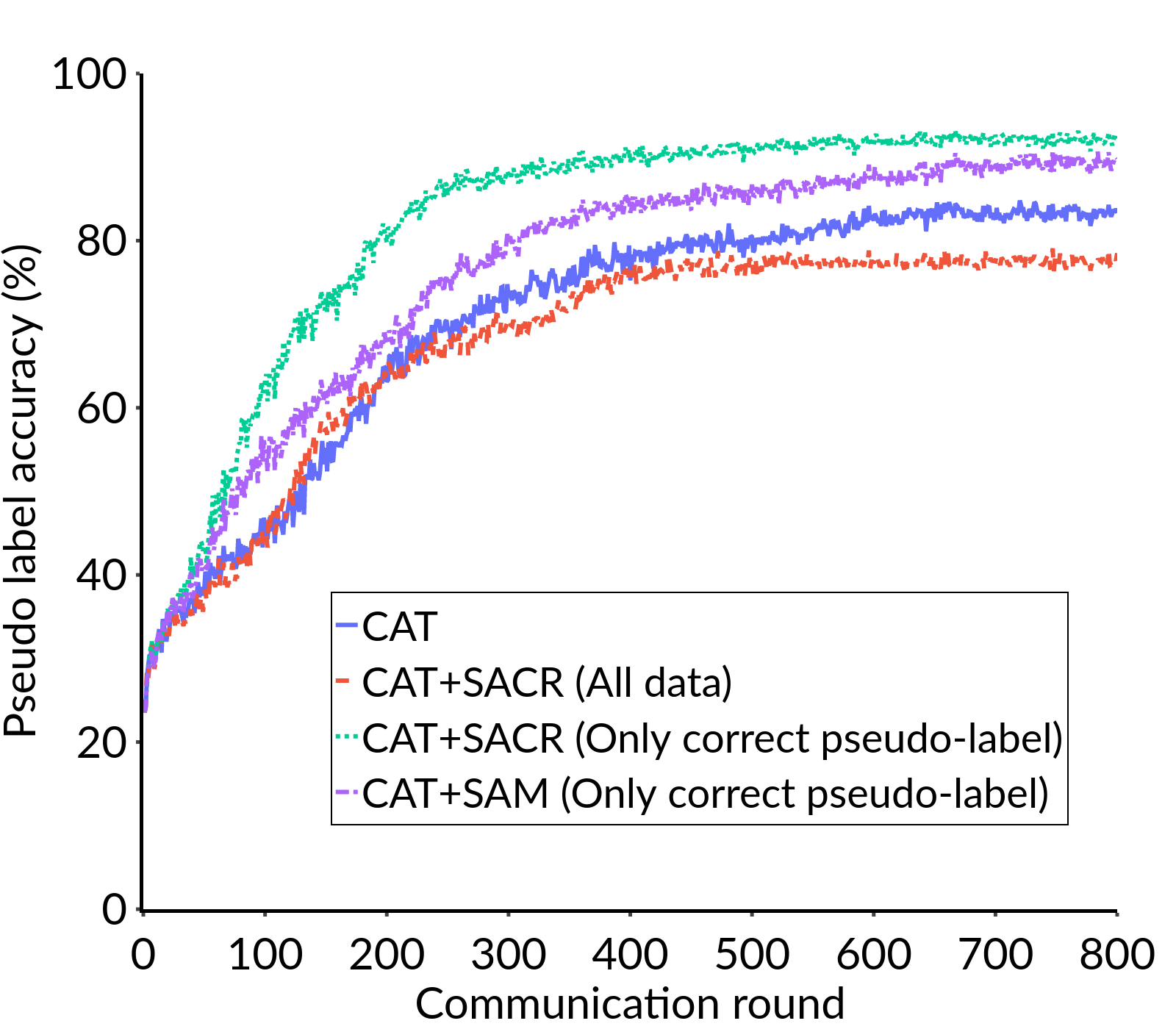}
        \caption{Pseudo-label accuracy.}
        \label{fig:wron-pseudo-2}
    \end{subfigure}
    \caption{Test accuracy and pseudo-label accuracy on the CIFAR10 dataset with 40 labels, balanced IID setting. Client-specific Adaptive Thresholding~(CAT) is used as the baseline. Applying Sharpness-aware Consistency Regularization~(SACR) to all data, including wrongly pseudo-labeled data, degrades performance than using only CAT, while applying SACR to correctly labeled data improves performance. SACR also outperforms the standard SAM objective (CAT+SAM).}
    \label{fig:wrong-pseudo}
    \vspace{-1em}
\end{figure}

\subsection{Impact of incorrect pseudo-labels on sharpness-aware consistency regularization}
\label{eval:wrong}
We investigate the impact of incorrect pseudo-labeled data on Sharpness-Aware Consistency Regularization~(SACR). We compare the performance of SACR in two scenarios: when applied only to correctly pseudo-labeled data assuming that we know the ground truth labels to assess the upper bound of SACR, and when applied to all pseudo-labeled data, including incorrectly pseudo-labeled samples. We examine when Client-specific Adaptive Thresholding~(CAT) is used in both scenarios.

Fig.~\ref{fig:wrong-pseudo} reports the test accuracy and pseudo-label accuracy for the following cases: CAT alone, CAT+SACR (all data), and CAT+SACR (only correct pseudo-labels). CAT+SACR (only correct pseudo-labels) achieves high pseudo-label accuracy, indicating that SACR can effectively reduce \textit{confirmation bias} when applied to correctly pseudo-labeled data. Conversely, when SACR is applied to all data, including wrongly pseudo-labeled samples, the performance significantly decreases and shows worse performance than using only CAT. This emphasizes the importance of applying SACR exclusively to carefully selected data samples that are highly likely to be correct.

\subsection{Comparison with the standard SAM objective}
\label{eval:sam}
We compare the proposed Sharpness-aware Consistency Regularization (SACR) to the standard Sharpness-Aware Minimization (SAM) objective. Both SAM and SACR perturb the model to maximize the given loss function. However, in SACR, the distance between the model outputs of perturbed and original model weights is minimized, while SAM takes the gradient of the given loss function at the perturbed weights.

Fig.~\ref{fig:wrong-pseudo} shows the test and pseudo-label accuracy using the standard SAM objective versus SACR. We examine the effects of SAM and SACR when applied only to correctly labeled samples in conjunction with Client-specific Adaptive Thresholding (CAT). Although SAM improves the performance over standalone CAT, SACR outperforms the standard SAM in convergence speed and final accuracy.
The effectiveness of SACR can be attributed to the fundamental differences between SAM and SACR. SAM \textit{explores} the given loss landscape in search of a flat local minima. In contrast, SACR \textit{changes} the loss landscape by explicitly incorporating an additional consistency regularization term.

\section{Discussion and conclusion}

We introduced a novel federated learning algorithm, \textbf{F}ew-\textbf{L}abels \textbf{F}ederated semi-supervised \textbf{L}earning, \putname{}, that addresses the challenge of few-labels settings in Federated Semi-Supervised Learning~(FSSL) for unlabeled clients. \putname{} effectively reduces the \textit{confirmation bias} through three key strategies: (1) \textit{client-specific adaptive thresholding}, which adjusts the pseudo-labeling threshold based on each client's learning status; (2) \textit{sharpness-aware consistency regularization}, which ensures consistency between the original and the adversarially perturbed models with carefully selected high-confidence pseudo labels; and (3) \textit{learning status-aware aggregation}, which incorporates each client's learning progress into the aggregation of client models. \putname{} closes the performance gap between SSL and FSSL, making FSSL an effective solution for practical scenarios.

\paragraph{Limitations and future work} Our approach introduces additional computational demands on clients, as \textit{client-specific adaptive thresholding} generates more pseudo-labels than traditional fixed threshold methods. Furthermore, \textit{sharpness-aware consistency regularization} adds an extra inference step with a perturbed model, increasing the computational burden. While our study is grounded in empirical findings, a promising future direction is to theoretically analyze the impact of the proposed methods, particularly in understanding how the generalization of incorrectly pseudo-labeled data affects overall performance.

\begin{ack}
This work was funded by the National Research Foundation of Korea (NRF), funded by the Ministry of Science and ICT (MSIT) under grant RS-2024-00464269 and the National Research Foundation of Korea (NRF) grant funded by the Korea government (MSIT) (RS-2024-00337007). ※ MSIT: Ministry of Science and ICT. 

\end{ack}

\bibliography{ref}
\bibliographystyle{unsrt}


\newpage
\appendix

\section*{\centering \Large (FL)${}^2$: Overcoming Few Labels in Federated Semi-Supervised Learning}
\vspace{-1em}
\section*{\centering Appendix}

\section{Algorithm}
\label{appendix:algorithm}
\begin{algorithm}
    \setstretch{1.35}
    \caption{ \putname{}: \textbf{F}ew-\textbf{L}abels \textbf{F}ederated semi-supervised \textbf{L}earning}
    \begin{algorithmic}[1]
        \State \textbf{Input:} Small labeled dataset $\mathcal{D}^S_L = \{(x_{b}, y_{b}) : b \in [N_{L}]\}$ at server. Unlabeled dataset $\mathcal{D}^m_U = \{u_{b} : b \in [N^{m}_U]\}, m \in [M]$ distributed over $M$ clients.  $\tau_f$ is fixed threshold. $B_c$ is client batch size. $\omega(\cdot)$ indicates weak data augmentation, and $\Omega(\cdot)$ indicates strong data augmentation. $\mathcal{H}(\cdot, \cdot)$ indicates cross-entropy loss. $\ell_d(\cdot , \cdot)$ is KL-divergence loss.
        \State \textbf{Initialize} global model weight $W^g_0$

        \For{each communication round $t$ }
            \State $W^g_t \leftarrow$ \textbf{ServerUpdate}$(W^g_t, \mathcal{D}^S_L)$ \Comment{Supervised server update with $D_L^S$}
            \State Update sBN statistics
            \State Server samples clients $K \in [M]$
            \State Server broadcasts $W_t^g$ to selected $K$ clients
            \For {each client $k \in [K]$} \textbf{parallel}
                \State $\tau_t^k \leftarrow \textrm{\textbf{AdaptiveThreshold}}(W_t^g, \mathcal{D}^k_U)$
                \State $W^k_t \leftarrow W^g_t$
                \For {each local step}
                    \State Sample $B_s$ sized batch $u_b$ from $\mathcal{D}^k_U$
                    \State $q_b^g \leftarrow p_{W_t^g}(y | \omega(u_b))$
                    \State $\hat{q}_b^g \leftarrow \mathrm{\textbf{OneHot}} (q_b^g)$
                    \State $Q_b \leftarrow p_{W_t^k}(y | \Omega(u_b))$
                    \State $\mathcal{L}_a^k \leftarrow \frac{1}{\mu B_c} \sum^{\mu B_c}_{b=1} \mathbbm{1} (\max(q^g_b) > \tau_{t}^{k}(\mathrm{arg max}(q^g_b)) \cdot \mathcal{H}(\hat{q}^g_b, Q_b)$
                    \State $\mathcal{L}_p^k \leftarrow \frac{1}{\mu B_c} \sum^{\mu B_c}_{b=1} \mathbbm{1} (\max(q_b^g) > \tau_f) \cdot \mathcal{H}(\hat{q}_b^g, Q_b)$
                    \State $W^{k*}_t \leftarrow W^k_t + \frac{\nabla_{W^k_t} \mathcal{L}_p^k}{\|\nabla_{W^k_t} \mathcal{L}_p^k\|_2}$
                     \State $Q_b^* \leftarrow p_{W_t^{k*}}(y | \Omega(u_b))$
                    \State $\mathcal{L}_{cs}^k \leftarrow \ell_d(Q_b^*, Q_b)$
                    \State $W^k_t \leftarrow W^k_t - \eta \nabla_{W^k_t} (w_{cs}\mathcal{L}_{cs}^k + w_a\mathcal{L}_a^k)$
                \EndFor
            \EndFor
            \State Clients uploads $W^k_t, \tau_t^k$ to server
            \State $\beta_k \leftarrow \frac{1-\tau_t^k}{\sum_{k=1}^K(1-\tau_t^k)}$
            \State $W^g_{t+1} \leftarrow \sum_{k=1}^K \beta_k W^k_t$
        \EndFor

    \end{algorithmic}
\end{algorithm}

\section{More experiment results}
\label{appendix:more-exp}
\begin{table}[h]
    \centering
    \caption{More evaluation results of \putname{} compared with SemiFL on Fashion-MNIST and AGNews dataset. We report the average accuracy(\%) and standard deviation across three runs with different random seeds.}
    \vspace{0.5em}
    \label{tab:add-result}
    \begin{adjustbox}{max width=1.0\textwidth}
    \begin{threeparttable}
        \begin{tabular}{cccc}
        \toprule
        \multicolumn{2}{c}{Dataset}                                                                     & Fashion-MNIST    & AGNews                                             \\ 
        \cmidrule(lr){3-3} \cmidrule(lr){4-4}
\multicolumn{2}{c}{\# of labeled data samples ($N_L$)}                                                  & 40             & 20 \\
        \midrule
        \multirow{2}{*}{\makecell{Unbalanced Non-IID,\\ Dir(0.3)}} 
                                                      & SemiFL                   & 12.8(4.8)     & 59.1(13.7)\\
                                                          & \putname{}           & \textbf{63.2(0.5)}    & \textbf{73.6(3.7)}\\
        \midrule
        \multirow{2}{*}{Balanced IID}             
                                                       & SemiFL                  & 10.0(0.0)   & 47.4(14.3) \\
                                                      & \putname{}               & \textbf{49.8(34.5)}\tnote{1}   & \textbf{87.0(0.6)} \\  
        \bottomrule
    
\end{tabular}
\begin{tablenotes}
\item[1] \small{One run failed, resulting in only 10\% accuracy, while the other two runs achieved accuracies of 69.0\% and 70.4\%}.
\end{tablenotes}
\end{threeparttable}
\end{adjustbox}
\end{table}

We conducted additional experiments on the Fashion-MNIST~\cite{xiao2017fashion} and AGNews~\cite{zhang2015character} datasets, and the result is shown in Table~\ref{tab:add-result}. For Fashion-MNIST, we used the WideResNet28x2 architecture, consistent with the SVHN and CIFAR-10 experiments. We compared \putname{} with the previous state-of-the-art, SemiFL. When trained with only 40 labeled samples, SemiFL failed in all three runs under the balanced IID setting and in two out of three runs under the non-IID-0.3 setting, resulting in accuracies around 10\%. In the single successful run under non-IID-0.3, SemiFL achieved an accuracy of 18.4\%. In contrast, \putname{} successfully trained in all three runs under non-IID-0.3 and in two out of three runs under balanced IID. In the one failed balanced IID run, the accuracy dropped to around 10\%, while in the successful runs, it reached 69\% and 70.4\%. On average, \putname{} achieved 63.2\% accuracy under the non-IID-0.3 setting, demonstrating its robustness and effectiveness even with minimal labeled data.

For the AGNews dataset, we randomly sampled 12,500 training samples per class from a total of 50,000 samples and applied back-translation for strong data augmentation, following the SoftMatch~\cite{chen2023softmatch} approach. We used the bert-base-uncased~\cite{devlin2019bertpretrainingdeepbidirectional} model as the backbone, freezing the BERT parameters and training only the linear classifier for 20 epochs. Since the mixup loss cannot be applied to NLP datasets, we compared \putname{} to SemiFL without the mixup loss. \putname{} significantly outperformed the baseline, with a 39.6\% accuracy improvement under the balanced IID setting and a 14.5\% improvement under the non-IID-0.3 setting. With only 20 labeled samples, SemiFL showed substantial performance variability, with standard deviations of 14.3 and 13.7 for the IID and non-IID-0.3 settings, respectively. In contrast, \putname{} delivered more consistent results, achieving standard deviations of 0.6 for IID and 3.7 for non-IID-0.3.

\section{Details of learning setup}
\label{appendix:learning}
All experimental results for FedMatch and FedCon were reproduced using the official PyTorch implementation of FedCon (\href{https://github.com/zewei-long/fedcon-pytorch}{zewei-long/fedcon-pytorch}), which is included in our repository. For SemiFL and \putname{} results, we implemented our own pipeline based on the GitHub repository for SemiFL (\href{https://github.com/diaoenmao/SemiFL-Semi-Supervised-Federated-Learning-for-Unlabeled-Clients-with-Alternate-Training}{diaoenmao/SemiFL-Semi-Supervised-Federated-Learning-for-Unlabeled-Clients-with-Alternate-Training}).

In Table~\ref{table:hyperparam}, we list the hyperparameters used in the experiments. We utilized SGD as our optimizer and a cosine learning rate decay as our scheduler. Additionally, we adapted the principles of adaptive federated optimization~\cite{fedopt} into our FedAvg algorithm by introducing a FedAvg optimizer. Instead of simply aggregating the local models' weights from clients and using this as the new global model's weights, as done in FedAvg, we calculated the difference between the aggregated local models' weights and the global model's weights. This difference was treated as the gradient of the global model's weights, which was then used to optimize the global model through the FedAvg optimizer. We set $\beta_l=0.9$ for the local optimizer and $\beta_g=0.5$ for the FedAvg optimizer.

For training with labeled data at the server, we used the standard supervised loss. For local training at unlabeled clients, our objective function was a weighted sum of the unsupervised loss, the fairness loss (from client-specific adaptive thresholding), and the consistency loss (from sharpness-aware regularization), with loss weights of $w_a=1$, and $w_{cs} =1$, respectively. For sharpness-aware regularization, we used $\rho=0.5$ and $\tau_f=0.95$. We also used an unlabeled batch size of 32, except for SemiFL, where training became unstable with this batch size, so we opted for a batch size of 10 as in the original paper.

\begin{table}[h]
\centering
\caption {Hyperparameters in our experiments}
\label{table:hyperparam}
\begin{adjustbox}{max width=0.9\textwidth}
\begin{tabular}{cc|cccc}
\hline
\multicolumn{2}{c|}{Method}                                              & \multicolumn{1}{l}{FedMatch~\cite{jeong2020federated}} & \multicolumn{1}{l}{FedCon~\cite{long2021fedcon}} & \multicolumn{1}{l}{SemiFL~\cite{diao2022semifl}} & \multicolumn{1}{l}{\putname{}} \\ \hline
\multicolumn{1}{c|}{\multirow{7}{*}{Server}}  & Batch size               & \multicolumn{4}{c}{10}                                                                                            \\ \cline{2-6} 
\multicolumn{1}{c|}{}                         & Epoch                    & \multicolumn{4}{c}{5}                                                                                             \\ \cline{2-6} 
\multicolumn{1}{c|}{}                         & Optimizer                & \multicolumn{4}{c}{SGD}                                                                                           \\ \cline{2-6} 
\multicolumn{1}{c|}{}                         & Learning rate            & \multicolumn{4}{c}{0.03}                                                                                          \\ \cline{2-6} 
\multicolumn{1}{c|}{}                         & Weight decay             & \multicolumn{4}{c}{0.0005}                                                                                        \\ \cline{2-6} 
\multicolumn{1}{c|}{}                         & Momentum                 & \multicolumn{4}{c}{0.9}                                                                                           \\ \cline{2-6} 
\multicolumn{1}{c|}{}                         & Nesterov                 & \multicolumn{4}{c}{\checkmark}                                                                                          \\ \hline 
\multicolumn{1}{c|}{\multirow{11}{*}{Client}}                         & Epoch                    & \multicolumn{4}{c}{5}                                                                                             \\ \cline{2-6} 
\multicolumn{1}{c|}{}                         & Optimizer                & \multicolumn{4}{c}{SGD}                                                                                           \\ \cline{2-6} 
\multicolumn{1}{c|}{}                         & Learning rate            & \multicolumn{4}{c}{0.03}                                                                                          \\ \cline{2-6} 
\multicolumn{1}{c|}{}                         & Weight decay             & \multicolumn{4}{c}{0.0005}                                                                                        \\ \cline{2-6} 
\multicolumn{1}{c|}{}                         & Momentum                 & \multicolumn{4}{c}{0.9}                                                                                           \\ \cline{2-6} 
\multicolumn{1}{c|}{}                         & Nesterov                 & \multicolumn{4}{c}{\checkmark}                                                                                              \\ \cline{2-6}
\multicolumn{1}{c|}{} & Batch size               & 32                           & \multicolumn{1}{|c}{32}                         & \multicolumn{1}{|c}{10}                         & \multicolumn{1}{|c}{32}                       \\ \cline{2-6}
\multicolumn{1}{c|}{}                         & Unsupervised loss weight~($w_a$) &            N/A                  &     \multicolumn{1}{|c}{N/A}                       &           \multicolumn{1}{|c}{N/A}                 & \multicolumn{1}{|c}{1.0}                      \\ \cline{2-6} 
\multicolumn{1}{c|}{}                         & Consistency loss weight~($w_{cs}$)  &             N/A                 &               \multicolumn{1}{|c}{N/A}             &             \multicolumn{1}{|c}{N/A}               & \multicolumn{1}{|c}{1.0}                      \\ \cline{2-6} 
\multicolumn{1}{c|}{}                         & $\rho$                      &              N/A                &            \multicolumn{1}{|c}{N/A}                &             \multicolumn{1}{|c}{N/A}               & \multicolumn{1}{|c}{0.1, 1.0}                      \\ \cline{2-6} 
\multicolumn{1}{c|}{}                         & $\tau_f$                   &             N/A                 &            \multicolumn{1}{|c}{N/A}                &               \multicolumn{1}{|c}{N/A}             & \multicolumn{1}{|c}{0.95}                     \\ \hline
\multicolumn{1}{c|}{\multirow{3}{*}{Global}}  & Communication round      & \multicolumn{4}{c}{800}                                                                                           \\ \cline{2-6} 
\multicolumn{1}{c|}{}                         & FedAvg momentum          & \multicolumn{4}{c}{0.5}                                                                                           \\ \cline{2-6} 
\multicolumn{1}{c|}{}                         & Scheduler                & \multicolumn{4}{c}{Cosine Annealing}                                                                              \\ \hline
\end{tabular}%
\end{adjustbox}
\end{table}


\section{Federated learning~(FL)}
\label{appendix:fl}
FL enables collaborative learning by sharing model updates while maintaining data privacy and distribution across clients. A widely used FL algorithm is FedAvg~\cite{fedavg}, which creates a global model by weighted-aggregating parameters from randomly selected clients, achieving convergence after multiple communication rounds. FedProx~\cite{fedprox} enhances the stability of FedAvg in non-IID settings by averaging local updates uniformly and incorporating proximal regularization against the global weights. FedOpt~\cite{fedopt} improves performance over FedAvg by introducing federated versions of adaptive optimizers. FedSim~\cite{fedsim} uses a similarity-guided approach, which clusters clients with similar gradients to enable local aggregations. However, most FL methods assume that labeled data is available to the client, which is impractical.

\section{Static batch normalization~(sBN)}
\label{appendix:sbn}
Following HeteroFL~\cite{diao2020heterofl} and SemiFL~\cite{diao2022semifl}, we adopt the Static Batch Normalization (sBN) into our client-weight aggregation algorithm at the server. This approach is specifically designed for federated learning (FL) settings and has been shown to accelerates convergence and improve the performance of the trained model compared to naive adoption of other normalization method for centralized setting such as Batch Normalization (BN)~\cite{batchnorm}, InstanceNorm~\cite{instancenorm}, GroupNorm~\cite{groupnorm}, and LayerNorm~\cite{layernorm}.

In detail, unlike the normal training phase where each client tracks its own running statistics and affine parameters of the BN layer to send to the server for aggregation, sBN disables the tracking of running statistics during local training at clients. At the beginning of each communication round, before local training begins, the server sequentially sends the model to all active clients. At each client, running statistics tracking is temporarily enabled (without momentum), and all training data is fed into the global model to cumulatively compute the mean and variance for the BN layers in the model.

\end{document}